\pdfoutput=1
\documentclass[11pt]{article}
\usepackage[preprint]{acl}

\usepackage{times}
\usepackage{latexsym}

\usepackage[T1]{fontenc}
\usepackage[utf8]{inputenc}

\usepackage{microtype}

\usepackage{inconsolata}

\usepackage{graphicx}
\usepackage{arabtex}
\usepackage{float} % for [H] specifier
\usepackage{geometry} % for adjusting margins
\usepackage{array} % for customizing table layout
\usepackage{tabularx} % for tables with fixed width columns
\usepackage{multirow}

\title{Nullpointer at ArAIEval Shared Task: \\Arabic Propagandist Technique Detection with Token-to-Word Mapping\\ in Sequence Tagging}

\author{
        Abrar Abir \\
        Computer Science Program, \\
        Carnegie Mellon University in Qatar\\
        Doha, Qatar\\
        \texttt{abir@cmu.edu} 
    \And 
        Kemal Oflazer\thanks{At the time of the research, Kemal Oflazer was also affiliated with Carnegie Mellon University, Qatar.} \\
        Language Technologies Institute, \\
        Carnegie Mellon University \\
        Pittsburgh, PA, USA\\
        \texttt{ko@andrew.cmu.edu}
        }

\begin{document}
\maketitle

 {\centerline{\large\bf Abstract}%
		  \begin{list}{}%
		     {\setlength{\rightmargin}{0.6cm}%
		      \setlength{\leftmargin}{0.6cm}}%
		   \item[]\ignorespaces%
     This paper investigates the optimization of propaganda technique detection in Arabic text, including tweets \& news paragraphs, from ArAIEval shared task 1. Our approach involves fine-tuning the AraBERT v2 model with a neural network classifier for sequence tagging.
Experimental results show relying on the first token of the word for technique prediction produces the best performance. In addition, incorporating genre information as a feature further enhances the model's performance. Our system achieved a score of 25.41, placing us 4$^{th}$ on the leaderboard. Subsequent post-submission improvements further raised  our score to 26.68.
%		   \@setsize\normalsize{12pt}\xpt\@xpt
		   %
		 \end{list}

\setcode{utf8}
\section{Introduction}
The spread of propagandistic content on social media and news platforms poses a significant challenge to the integrity of information consumed by the public. Detecting and countering such content is crucial for maintaining a well-informed society. Identifying propaganda techniques in Arabic texts presents unique challenges due to the language's complex morphology and the use of diverse dialects.
Our main contribution is developing an optimized system that leverages AraBERT for embeddings and a neural network classifier for sequence tagging to detect propagandistic techniques in Arabic texts. We propose a robust preprocessing pipeline for Arabic, addressing Unicode inconsistencies, misaligned annotations, and user-mention normalization. We also evaluate and compare different sequence tagging approaches.

The paper is structured as follows: Section 2 describes the related work. Section 3 describes the data preprocessing steps. Section 4 discusses the sequence tagging prediction approaches. Section 5 presents the details of our system. Section 6 reports on the experimental results and discussions. Section 7 concludes the paper.

\section{Related Work}
Detecting propaganda and persuasion techniques in Arabic text has gained significant attention in recent years. Hasanain \textit{et al.} \shortcite{araieval:arabicnlp2023-overview} have introduced the ArAIEval shared task, focusing on detecting propagandistic techniques in unimodal and multimodal Arabic content, highlighting the importance of this task in the Arabic language context.
Alam \textit{et al.} \shortcite{alam-etal-2022-overview} have provided an overview of the WANLP 2022 shared task on propaganda detection in Arabic, discussing the challenges and significance of this task in the Arabic language domain.
Several studies have explored the use of large language models for propaganda detection. Hasanain \textit{et al.} \shortcite{hasanain2023large} investigated the application of large language models for propaganda span annotation, showcasing their potential for identifying propagandistic content. In a follow-up study, Hasanain \textit{et al.} \shortcite{hasanain2024can} have evaluated the capability of GPT-4 in identifying propaganda and annotating propaganda spans in news articles, demonstrating the effectiveness of advanced language models in this domain.

The ArAIEval shared task, organized by Hasanain \textit{et al.} \shortcite{araieval:arabicnlp2023-overview}, focused on persuasion techniques and disinformation detection in Arabic text, emphasizing the need for language-specific approaches in tackling these issues.
These studies demonstrate the growing interest in propaganda detection and the diverse approaches explored to address this challenge, particularly in Arabic. Our work builds upon these existing efforts, focusing on the role of token-to-word mapping in Arabic sequence tagging for propaganda detection.

% \section{Data}

\section{Preprocessing the Data}
In this section, we describe our data and the preprocessing steps applied to ensure the annotation consistency and accuracy in the text snippets. %The preprocessing involves handling special Unicode characters, misaligned span annotations, and inconsistencies in user mentions.
\paragraph{Dataset} The dataset is taken from the ArAIEval 2024 task 1.\footnote{https://gitlab.com/araieval/araieval\_arabicnlp24/-/tree/main/task1} The dataset contains JSONL files with UTF-8 text encodings. Given a multigenre text snippet (a news paragraph or a tweet) from the dataset, the task is to detect the propaganda techniques used in the text together with the exact span(s) in which each propaganda technique appears. Table \ref{tab:data-distribution} presents basic statistics on the dataset.

% \textbf{You should describe what the data set is and whether it is available publicly and where, etc.}}

\paragraph{Removal of Unicode Control Characters}

To handle emojis and other Unicode characters commonly found in tweets, we remove all Unicode characters classified as control (``Cf'') or other (``Co'') and replace them with spaces. Replacement with spaces instead of just removing them preserves the character positions and avoids disrupting the annotation spans in the text.

\paragraph{Handling Misaligned Span Annotations}
For some annotations in the dataset, the reported span of the propagandistic technique ("start" and "end") does not match with the actual start and character positions of the propagandistic technique substring in the main string; e.g., 
% (\textbf{I do not understand what you mean here)}. 
% \textbf{scale the jpg a bit more here.  You can get rid of the useless gray space on the left and right, which will make the figure look larger}

\begin{figure}[h]
  \includegraphics[width=\columnwidth]{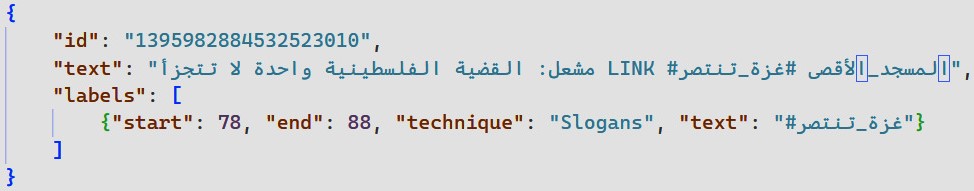}
\end{figure}

\noindent
In this instance, the annotation span starts at character 78 and ends at character 88, but the text is only 69 characters long. To address this issue, we completely ignore the reported span ("start" and "end") and perform a substring search using the text technique provided in the annotation to determine the correct character spans. %This approach ensures that the annotations accurately reflect the positions of the techniques in the text.

\paragraph{Normalizing User Mentions}

In the dataset, user mentions in the text were replaced with a generic placeholder ``@USER''. However, this replacement was not reflected in the technique annotations. Consider the following example:\\  
% \textbf{Remove the wasted space on the left in the figure -- the row numbers}
% \begin{figure}[h]
\includegraphics[width=\columnwidth]{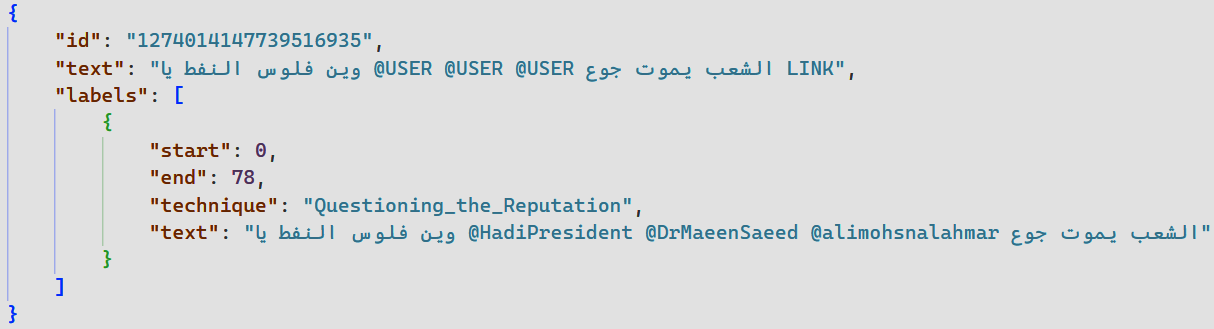}
% \end{figure}

We replaced all specific user mentions in the technique annotations to resolve this discrepancy with the generic ``@USER'' placeholder. This normalization ensured consistency between the text and its annotations.

\paragraph{Manual Handling of Failed Substring Searches}

Despite the automated preprocessing steps, there were cases where the substring search failed due to discrepancies in Unicode handling. For instance:\\
% \textbf{Remove the wasted space on the left in the figure -- the row numbers}
% \begin{figure}[h]
\includegraphics[width=\columnwidth]{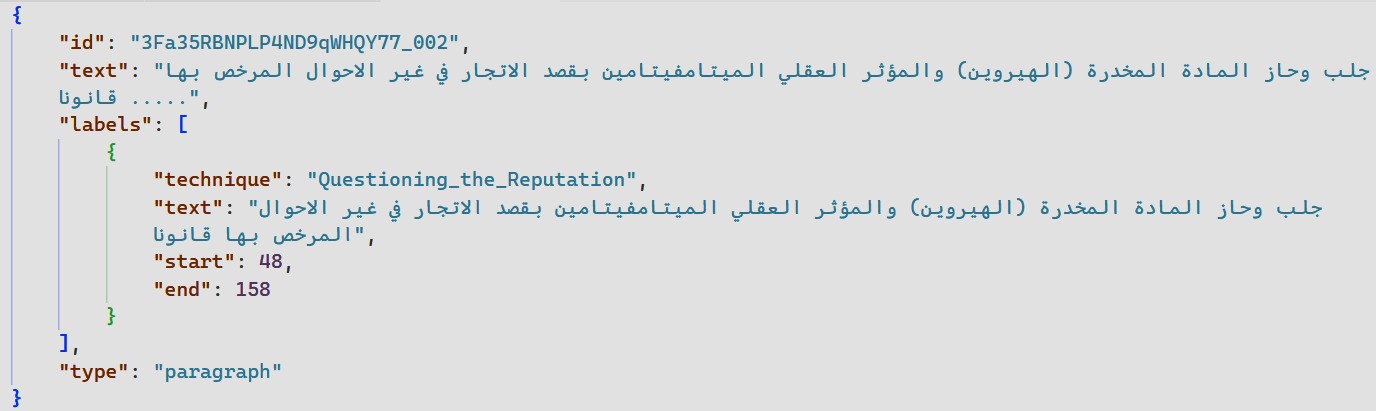}
% \end{figure}
Here, the parentheses around \mbox{\RL{الهيروين}} \textit{[alhirwin - Heroin]} were reversed in the main text compared to the annotation, likely due to improper Unicode handling of [U+202C] and [U+202D]. Such cases were manually corrected to ensure the annotations matched the text accurately.

These preprocessing steps ensured the dataset was clean and the annotations aligned correctly with the text.\\%, providing a solid foundation for the subsequent analysis and modeling tasks.\\

%\begin{table}[!ht]
%\centering
%\small
%\caption{Distribution of Propagandistic Techniques }
% \label{tab:propagandistic-techniques}
%\begin{tabular}{lrrr}
%\hline
%\textbf{Propagandistic} & \textbf{Train} & \textbf{Dev} & \textbf{Test} \\
%\textbf{Technique} & \textbf{(6997)} & \textbf{(921)} & \textbf{(1046)} \\
%\hline
%Loaded\_Language & 8779 & 1073 & 1429 \\
%Name\_Calling-Labeling & 2243 & 330 & 432 \\
%Exaggeration-Minimisation & 1077 & 141 & 133 \\
%Doubt & 293 & 44 & 26 \\
%Causal\_Oversimplification & 310 & 38 & 21 \\
%Obfuscation-Vagueness & 578 & 67 & 55 \\
%-Confusion\\
%Appeal\_to\_Fear-Prejudice & 180 & 32 & 7 \\
%Slogans & 190 & 37 & 7 \\
%Appeal\_to\_Authority & 218 & 28 & 13 \\
%Flag\_Waving & 218 & 34 & 17 \\
%Repetition & 144 & 17 & 17 \\
%Questioning\_the\_Reputation & 872 & 128 & 106 \\
%Appeal\_to\_Values & 116 & 24 & 5 \\
%Appeal\_to\_Hypocrisy & 108 & 16 & 7 \\
%Consequential & 83 & 11 & 6 \\
%Oversimplification\\
%Conversation\_Killer & 69 & 10 & 6 \\
%Appeal\_to\_Time & 54 & 6 & 8 \\
%Appeal\_to\_Popularity & 46 & 4 & 3 \\
%Red\_Herring & 41 & 5 & 3 \\
%False\_Dilemma-No\_Choice & 74 & 9 & 5 \\
%Guilt\_by\_Association & 23 & 2 & 4 \\
%Straw\_Man & 23 & 3 & 2 \\
%Whataboutism & 26 & 5 & 1 \\
%\hline
%\textbf{Total} & \textbf{15765} & \textbf{2064} & \textbf{2313}\\
%\end{tabular}
%\end{table}

\vspace{-0.5cm} % reduce space between tables

\begin{table}[t]
\centering
\caption{Data Distribution by Type}
\label{tab:data-distribution}
\begin{tabular}{lcccc}
\hline
Type & Train & Dev & Test & \textbf{Total}\\
\hline
Tweet & 995 & 249 & 260 & 1504 \\
Paragraph & 6002 & 672 & 786 & 7476\\
\hline
\end{tabular}

\end{table}

\section{Sequence Tagging Prediction Approaches}

This section discusses various approaches to sequence tagging for detecting propagandistic techniques in text. 
% The task involves labeling each token or word in a text snippet and subsequently identifying spans corresponding to different propagandistic techniques. We explore three main approaches: token-level training and prediction, token-level training with word-level prediction, and word-level training and prediction.

\subsection{Token-Level Training and Prediction}

In this approach, the model is trained on labels assigned to each token, and predictions are made for each token. The predicted spans of propagandistic techniques are then computed based on these token-level predictions. One limitation of this method is that the predicted spans might start or end in the middle of a word since words are often broken into multiple tokens.

\subsection{Token-Level Training with Word-Level Prediction}

To address the issue of token-level span inaccuracies, we explore two methods where the model is still trained on token-level labels but predictions are aggregated at the word level.

\paragraph{Majority Label Assignment}
In this approach, each word, when broken into multiple tokens, receives multiple labels. We assign the word the label that appears most frequently among its tokens. For instance, if a word is split into three tokens and one token is predicted to have label 1 while the other two tokens are predicted to have label 2, we assign the word label 2. %This majority voting approach helps in consolidating token-level predictions into a coherent word-level label.

\paragraph{First Token Label Assignment} In this approach, for each word broken into multiple tokens, we only consider the label of the first token to assign the label to the entire word. %This method simplifies the label assignment process by relying on the prediction of the first token, which can be particularly effective if the first token captures the primary semantic role of the word.

\subsection{Word-Level Training and Prediction}

In this approach, the model is trained directly on labels assigned to each word rather than each token. To achieve this, we determine the embedding of each word by max-pooling the embeddings of its constituent tokens. This aggregated word embedding is then used for both training and prediction at the word level. %By focusing on word-level representations, this method ensures that the predicted spans align with actual word boundaries.

\section{The Propagandistic Technique Detection System}

In this section, we describe the system used for the task of detecting propagandistic techniques in Arabic text snippets. Our system is based on the AraBERT version 2 pre-trained model, followed by a neural network classifier for sequence tagging as shown in Figure \ref{fig:enter-label}. Figure \ref{fig:enter-label} (a) shows the structure of one of our classifiers and Figure \ref{fig:enter-label} (b) shows the overall structure. There is one classifier for each token in the input.

\begin{figure}[t]
    \centering
    \includegraphics[width=\columnwidth]{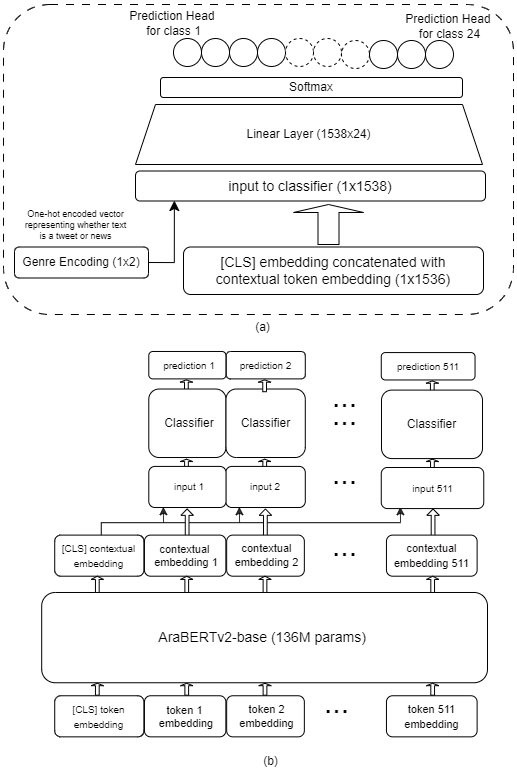}
    \caption{Diagram of the (a) Classifier, (b) Sequence Tagging System}
    \label{fig:enter-label}
\end{figure}
\paragraph{Embedding Generation \& Classification} For generating embeddings, we utilized the AraBERT version 2 base pre-trained model,\footnote{https://huggingface.co/aubmindlab/bert-base-arabert} a BERT-based model specifically designed for Arabic language processing. For each token, we concatenate the embedding of the \texttt{[CLS]} token with the token's contextual embeddings. Then, input is fed into a classifier to predict the labels corresponding to different propagandistic techniques. The architecture of the neural network is quite simple; it has no hidden layers, and  has a total of $1536 \cdot 24 = 36,864$ parameters to train.% The hyperparameters for the initial experiments were kept constant.

\paragraph{Sequence Tagging Approaches} We experimented with four different sequence tagging prediction approaches, as described in Section 3. %To ensure a fair comparison, we used the same neural network architecture and hyperparameters across all approaches. 
% The approaches compared were:

% \begin{itemize}
%     \item \textbf{Token-Level Training and Prediction:} Training and predicting at the token level.
%     \item \textbf{Token-Level Training with Word-Level Prediction:} Aggregating token-level predictions at the word level using majority label assignment and first token label assignment.
%     \item \textbf{Word-Level Training and Prediction:} Training and predicting at the word level using max-pooled embeddings of tokens.
% \end{itemize}

% After conducting these experiments, we selected the approach that yielded the best performance based on our evaluation metrics.
\paragraph{Encoding Genre as a Feature} We incorporated genre information as an additional feature in our model. Given the dataset comprises two distinct genres—tweets and news paragraphs—we encoded this information using a one-hot vector representation.
% Specifically, we created a one-hot vector of length 2, where each genre is represented by a unique binary pattern.\\
This two-dimensional genre vector was then concatenated with the ($768\cdot 2 = 1536$) two-dimensional embedding vector as the input to our neural network classifier. 
% Thus, we aimed to provide the model with contextual cues that could help improve the identification of propagandistic techniques, given that the style and structure of tweets differ from those of news paragraphs. 
% This approach allowed the model to leverage both the semantic content of the text and the genre-specific characteristics, potentially enhancing its predictive performance.

\paragraph{Hyperparameter Tuning} We experimented with various learning rates, batch sizes, number of epochs, and the inclusion of hidden layers. The hyperparameter tuning was performed using grid search and cross-validation to identify the optimal set of parameters for our final model.

\paragraph{Final Model Training} The final model was trained on concatenating the train and development dataset using the best sequence tagging approach and the optimal hyperparameters determined from our tuning experiments. This model was then used to predict the spans of propagandistic techniques in the test dataset, providing the results for our evaluation.

\section{Results and Discussions}
Table \ref{table:performance_combined} summarizes our results for different techniques and whether genre information is used or not. 
\begin{table}[ht]
\centering
% \tiny
\small
\caption{Performance Metrics (Micro F1 score) on Test File for Different Sequence Tagging Approaches - with \& without genre encoding}
\label{table:performance_combined}
\begin{tabular}{lcc}
\hline 
 & M1 score & M1 score\\
Approach & \textbf{with} Genre& \textbf{without} Genre\\
 % & Genre & Genre  \\
\hline 
Token-to-Token & 24.34 & 22.62\\\hline
Token-to-Word & \multirow{2}{2em}{20.73} & \multirow{2}{2em}{16.57}\\
(Majority-Label) & & \\\hline
Token-to-Word & \multirow{2}{2em}{\textbf{26.68}} & \multirow{2}{2em}{\textbf{24.37}}\\
(First-Label) &  & \\\hline
Word-to-Word & 12.94 & 13.22\\
\hline
\end{tabular}

\end{table}
\paragraph{\texttt{[CLS]} Token Embedding}  The \texttt{[CLS]} token serves as a representation of the entire input sequence. By concatenating this sequence representation with individual token embeddings, the model gains access to global and local contextual cues, enabling better predictions of sequence labels.  Simply adding the \texttt{[CLS]} embedding vector to the token embeddings produced slightly better outcomes compared to ignoring the \texttt{[CLS]} token altogether. However, this approach still fell short of the performance achieved through concatenation.
\paragraph{Different sequence tagging approaches} The word-level-first sequence tagging technique emerged as the most effective method. %, outperforming other approaches %in accurately identifying propagandistic techniques.
This result is consistent with the nature of the Arabic language, where the essence or primary meaning of a word is often captured by the first token, especially for verbs. %Unlike English, where word boundaries are clearer and each word typically represents a distinct concept, 
%Arabic words can convey complex meanings that may span multiple tokens. 
%By focusing on the first token, which encapsulates the core semantic content of the word, the word-level-first approach effectively captures the essence of the text snippet, leading to more accurate predictions of propagandistic techniques.
Conversely, the word-level-majority sequence tagging technique performed poorly compared to other methods. Despite aggregating token-level predictions at the word level, this approach did not yield significant improvements in accuracy. This suggests that the majority label assignment strategy may dilute the relevance of the primary propagandistic technique within each word, particularly in Arabic.%, where the first token often carries the most salient information.

% Interestingly, the token-level sequence tagging approach performed better than the word-level exact method. While both techniques operate at the token level, the token-level approach predicts labels for each individual token, allowing for more granular analysis of the text. This finer granularity may contribute to the improved performance by capturing subtle nuances and variations in propagandistic techniques within the text.

% Moreover, the observation that all techniques achieved their optimal test M1 score on epoch 3 of training highlights the efficiency of model convergence in the Arabic language context. This rapid convergence suggests that the models quickly learn the relevant patterns in the data, enabling them to achieve maximum accuracy with fewer training epochs.

\vspace*{-0.3cm}

\paragraph{Genre Encoding} Incorporating genre information into our model significantly improved performance across all sequence tagging approaches - except for the word-to-word approach. By providing the model with contextual cues about whether a text snippet was a tweet or a news paragraph, the model could better differentiate and identify the propagandistic techniques prevalent in each genre. This enhancement in performance, however, came at the cost of increased convergence time. Specifically, the inclusion of genre information extended the convergence period to around 10 epochs (as opposed to around 3 epochs ). %This suggests that while genre information adds valuable context, it also introduces additional complexity that the model needs more time to learn effectively. 
\paragraph{Classifier Architecture} In our experimentation, we explored various neural network architectures, ranging from models with 1 to 2 hidden layers containing 32 to 128 neurons on each layer to more complex configurations. Surprisingly, the best-performing model emerged as one with no hidden layers at all.
\section{Conclusion}
The study underscores the importance of methodological choices in improving the accuracy of propaganda technique detection models. Concatenating the \texttt{[CLS]} token embedding and leveraging the word-level-first sequence tagging approach, particularly when supplemented with genre information, proves to be a robust strategy. Future research should continue to explore the integration of additional contextual information and refine these techniques to further enhance the detection and understanding of propagandistic content across various languages and genres.

% \textbf{Use \{A\}rabic in the bib file.}  
% HOw do you know if the Last paper is actually accepted to the conference?}
\nocite{*}
\bibliography{custom}

\begin{thebibliography}{6}
\providecommand{\natexlab}[1]{#1}

\bibitem[{Alam et~al.(2024)Alam, Hasnat, Ahmed, Hasan, and
  Hasanain}]{alam2024armeme}
Firoj Alam, Abul Hasnat, Fatema Ahmed, Md~Arid Hasan, and Maram Hasanain. 2024.
\newblock \href {https://arxiv.org/abs/2406.03916} {{ArMeme}: Propagandistic
  content in {A}rabic memes}.
\newblock \emph{arXiv preprint arXiv:2406.03916}.

\bibitem[{Alam et~al.(2022)Alam, Mubarak, Zaghouani, Da~San~Martino, and
  Nakov}]{alam-etal-2022-overview}
Firoj Alam, Hamdy Mubarak, Wajdi Zaghouani, Giovanni Da~San~Martino, and
  Preslav Nakov. 2022.
\newblock \href {https://aclanthology.org/2022.wanlp-1.11} {Overview of the
  {WANLP} 2022 shared task on propaganda detection in {A}rabic}.
\newblock In \emph{Proceedings of the The Seventh Arabic Natural Language
  Processing Workshop (WANLP)}, pages 108--118, Abu Dhabi, United Arab Emirates
  (Hybrid). Association for Computational Linguistics.

\bibitem[{Hasanain et~al.(2023{\natexlab{a}})Hasanain, Ahmed, and
  Alam}]{hasanain2023large}
Maram Hasanain, Fatema Ahmed, and Firoj Alam. 2023{\natexlab{a}}.
\newblock Large language models for propaganda span annotation.
\newblock \emph{arXiv preprint arXiv:2311.09812}.

\bibitem[{Hasanain et~al.(2024{\natexlab{a}})Hasanain, Ahmed, and
  Alam}]{hasanain2024can}
Maram Hasanain, Fatema Ahmed, and Firoj Alam. 2024{\natexlab{a}}.
\newblock Can {GPT-4} identify propaganda? annotation and detection of
  propaganda spans in news articles.
\newblock In \emph{Proceedings of the 2024 Joint International Conference on
  Computational Linguistics, Language Resources And Evaluation, LREC-COLING
  2024}, Torino, Italy.

\bibitem[{Hasanain et~al.(2023{\natexlab{b}})Hasanain, Alam, Mubarak,
  Abdaljalil, Zaghouani, Nakov, {Da San Martino}, and
  Freihat}]{araieval:arabicnlp2023-overview}
Maram Hasanain, Firoj Alam, Hamdy Mubarak, Samir Abdaljalil, Wajdi Zaghouani,
  Preslav Nakov, Giovanni {Da San Martino}, and {Abed Alhakim} Freihat.
  2023{\natexlab{b}}.
\newblock {ArAIEval Shared Task}: Persuasion techniques and disinformation
  detection in {A}rabic text.
\newblock In \emph{Proceedings of the First Arabic Natural Language Processing
  Conference (ArabicNLP 2023)}, Singapore. Association for Computational
  Linguistics.

\bibitem[{Hasanain et~al.(2024{\natexlab{b}})Hasanain, Hasan, Ahmed, Suwaileh,
  Biswas, Zaghouani, and Alam}]{araieval:arabicnlp2024-overview}
Maram Hasanain, Md.~Arid Hasan, Fatema Ahmed, Reem Suwaileh, Md.~Rafiul Biswas,
  Wajdi Zaghouani, and Firoj Alam. 2024{\natexlab{b}}.
\newblock {ArAIEval Shared Task}: Propagandistic techniques detection in
  unimodal and multimodal {A}rabic content.
\newblock In \emph{Proceedings of the Second Arabic Natural Language Processing
  Conference (ArabicNLP 2024)}, Bangkok. Association for Computational
  Linguistics.

\end{thebibliography}
\end{document}